%% file: short4arxive.tex
\input{cscsstyle}
\usepackage[T1]{fontenc}
\usepackage[hyphens]{url}
\usepackage{hyperref}
\usepackage{textcomp}
\usepackage{xcolor}
\usepackage{graphicx}
\usepackage{authblk}

\begin{document}

\Title{Diffusion-Based Feature Denoising with NNMF for Robust handwritten digit multi-class classification}
\Author{Hiba Adil Al-kharsan$^{1}$, R\'{o}bert Rajk\'{o}$^{2}$ \\
\normalfont{\footnotesize{$^{1}$Doctoral School of Computer Science, University of Szeged, \'{A}rp\'{a}d t\'{e}r 2, H-6720 Szeged, Hungary; hibaadil@inf.u-szeged.hu\\
	$^{2}$University Research and Innovation Center (EKIK), \'{O}buda University, B\'{e}csi \'{u}t 96/b, H-1034 Budapest, Hungary; rajko.robert@uni-obuda.hu}}}

\begin{Abstract}
This work presents a robust multi-class classification framework for handwritten digits that combines diffusion-driven feature denoising with a hybrid feature representation. Inspired by our previous work on brain tumor classification~\cite{alkharsan2026diffusionbasedfeaturedenoisingusing}, the proposed approach operates in a feature space to improve the robustness to noise and adversarial attacks.
{This manuscript is submitted as an extended abstract rather than a full-length press-ready paper.}

First, the input images are converted into tight, interpretable exemplification using Non-negative Matrix Factorization (NNMF). In parallel, special deep features are extracted using a computational neural network (CNN). These integral features are combined into a united hybrid representation.
{The main objective of this work is to extend our previously validated two-class framework to a multi-class handwritten digit classification scenario.}

To improve robustness, a step diffusion operation is used in the feature space by gradually adding Gaussian noise. A feature denoiser network is trained to reverse this operation and rebuild clean representations from tilted inputs. The courteous features are then applied for multi-class classification.

The suggested method is evaluated in both baseline and adversarial settings using AutoAttack. The experimental outcome present that the diffusion-based hybrid model is both effective and robust, the CNN baseline models outperforming while maintain powerful classification performance. These results explain the activity of feature-level diffusion defense for reliable multi-class handwritten digit classification.


\end{Abstract}

\section{Introduction}

Handwritten digit multi-class classification is a major problem in machine learning and is widely used to evaluate classification models. Convolutional Neural Networks (CNNs) achieve high accuracy by learning feature representations directly from the data \cite{cnn}.
{Although handwritten digit classification is a classical and well-studied task, it is used here as a controlled benchmark to evaluate robustness-oriented feature-level defense mechanisms.}

However, deep models remain vulnerable to adversarial perturbations, in which small input modifications can significantly degrade performance \cite{adv}. To ensure a reliable evaluation, standardized structures such as AutoAttack are used primarily to estimate robustness \cite{autoattack}.

Feature extraction methods such as Non-negative Matrix Factorization (NNMF) provide a consolidated, interpretable representation, complemented by the distinctive power of CNN features \cite{nnmf}. In addition, diffusion-based models have been used to rebuild clean data from noisy information and improve robustness to perturbations \cite{diffusion01}
{NNMF does not directly operate well on mixed-set data; therefore, digit samples are handled in a category-aware manner before constructing the final hybrid representation.}
{Unlike standard handwritten digit classification studies, this work focuses on robustness-oriented feature-space processing and adversarial resistance rather than improving conventional classification accuracy.}

In this model, a handwritten digit data set provided as part of the built-in MATLAB examples is used for training and evaluation. 
{This dataset is selected for reproducibility and direct integration within the MATLAB-based implementation pipeline, and serves as a practical proxy similar to MNIST.}
We propose a hybrid framework that combines NNMF and CNN features with diffusion-based feature denoising to achieve robust multi-class classification. The suggested method is evaluated in both clean and adversarial tuning, explaining improved accuracy and robustness.

{The main contribution of this work is the integration of NNMF-based interpretable features, CNN-based deep features, and diffusion-based feature denoising into a unified pipeline evaluated under adversarial settings.}
{In contrast to our previous work, which focused on a two-class scenario, this study investigates the scalability and behavior of the proposed method in a multi-class setting.}

\section{Materials and Methods}
The suggested method is organized into five main stages: data set preparation, feature extraction, hybrid feature structure and classifier training, diffusion-based feature denoising, and robustness evaluation. These stages are intended to enhance the stability of the classification and the robustness of the adversaries.

{Due to the limited length of the extended abstract format, only the essential methodological components are summarized.}

\textbf{Preparation of the data set:}
In this stage, a handwritten digit data set is provided in MATLAB examples \cite{dataset} and used for training and evaluation. The data set comprises approximately 10,000 grayscale images in 10 classes corresponding to digits (0--9), with almost equal numbers of samples per class. Whole images are re-sized to a steady dimension to guaranty matching for the next processing steps.
The data set is collected into three subsets: 70\% (7,000 images) for training, 15\% (1,500 images) for validation, and 15\% (1,500 images) for testing. This stable division ensures that sufficient data is available for learning while maintaining reliable validation during training and unbiased evaluation on unseen samples.

This stage established the foundation of the proposed framework, ensuring that the input data is properly organized, normalized, and prepared for feature extraction and model training in the subsequent stages.

\textbf{Hybrid Feature Construction Using CNN and NNMF:}
 A CNN is trained in the images to extract  special features from a fully connected medium layer, which holds high-level spatial information \cite{li2021deep}.

In parallel, NNMF uses vectorized images to gain interpretable parts-based representations. The number of components is set to \(k = 30\), that is, a summarized feature space while preserving primary structural patterns \cite{gillis2020nonnegative}. The NNMF model is trained on the training set and applied to project the validation and test data.
Both CNN and NNMF features are normalized and concatenated to form a united hybrid feature vector. A lightweight neural network multi-class classifier is then trained on the combine features for multi-class classification. This hybrid representation improves performance over individual feature types by merging 
 and interpretable characteristics.
{The NNMF-based representation follows the principle validated in our previous two-class experiments, here extended to the multi-class case.}

\textbf{Diffusion-Based Feature Denoising:}
To improve robustness to noise and trouble, a diffusion-based denoising process is used in the feature space. The hybrid feature vectors gained from the previous stage are gradually corrupted by adding Gaussian noise at multiple steps, simulating degraded representations. 
A denoising network is then trained to reconstruct the original clean features from noisy input by learning the inverse diffusion process. This approach enables the model to recover stable and informative representations even under corrupted conditions.
 Using a feature space rather than an image space, the proposed method decreases computational complexity while preserving primary special information. The diffusion-based denoising mechanism improves the robustness of the hybrid model and improves the classification performance, following the principles of probabilistic diffusion modeling \cite{diffusion01}.
{The purpose of this stage is to investigate robustness improvement rather than maximizing clean-data accuracy.}

 \textbf{Adversarial Evaluation and Performance Analysis.}
In this stage, the trained data are exported to the ONNX format and evaluated using Python.. The exported classifier and denoising system are loaded and converted into PyTorch format for active deduction on both CPU and GPU environments.

The estimate is applied to clean and adversarial samples. Firstly, the clean test set is passed through both the baseline classifier and the defended model, which combines the diffusion-based denoising procedure. The defended model executes multiple stochastic forward passes and mediates the predictions to gain a stable outcome.

To estimate robustness, adversarial symbols are created using the AutoAttack framework in the \(L_{\infty}\) threat system with a fixed trouble budget. Specifically, parameter-free attacks such as APGD-CE and APGD-DLR are applied to produce strong adversarial samples \cite{autoattack}. These attacks are designed to provide a reliable and unified evaluation of the system's robustness.

The execution of both baseline and defended data is ranked using multiple metrics, including accuracy, precision, recall, F1-score, Matthews correlation coefficient (MCC), and balanced accuracy. In addition, probabilistic metrics such as ROC-AUC, log-loss, and Brier score are calculated to assess prediction confidence and calibration \cite {medium2023metric}.
{A direct comparison with state-of-the-art digit classification models is outside the scope of this extended abstract, as the focus is on robustness evaluation rather than absolute accuracy performance.}

The comparison between clean and adversarial results explains that the proposed diffusion-based defense significantly progresses robustness while maintaining competitive performance on baseline data. This emphasizes the efficiency of combining feature-level denoising with hybrid feature representations for reliable classification under adversarial case.

\section{Results and Discussion}

{The results should be interpreted in the context of robustness evaluation and methodological validation.}
The suggested system is estimated on clean and adversarial test data. Four rating scenarios are considered: the baseline model for clean data, the defended model for clean data, the baseline model on adversarial attack, and the defended model under adversarial attacks. Table~\ref{tab:results_full} shows performance comparison on clean and adversarial data with fixed $k = 30$, and Table~\ref{tab:results_full_optimized} shows the same metrics using the optimized $k = 70$ component number for NNMF. We can see a significant improvement on metrics after optimization for the case of Robust\_Def.  

\begin{table*}[t]
\centering
\resizebox{\textwidth}{!}{
\begin{tabular}{lccccccccc}
\hline
\textbf{Case} & \textbf{Accuracy} & \textbf{Precision} & \textbf{Recall} & \textbf{F1} & \textbf{MCC} & \textbf{Balanced Acc} & \textbf{ROC-AUC} & \textbf{LogLoss} & \textbf{Brier} \\
\hline
Clean\_Base  & 0.992 & 0.9921 & 0.992 & 0.9920 & 0.9911 & 0.992 & 0.99997 & 1.4756 & 0.6605 \\
Clean\_Def   & 0.942 & 0.9430 & 0.942 & 0.9418 & 0.9357 & 0.942 & 0.99827 & 1.5407 & 0.6822 \\
Robust\_Base & 0.507 & 0.7088 & 0.507 & 0.4869 & 0.4771 & 0.507 & 0.97116 & 1.8536 & 0.7740 \\
Robust\_Def  & 0.582 & 0.6333 & 0.582 & 0.5751 & 0.5408 & 0.582 & 0.96883 & 1.7957 & 0.7580 \\
\hline
\end{tabular}
}
\caption{Performance comparison on clean and adversarial data.}
\label{tab:results_full}
\end{table*}

\begin{table*}[t]
	\centering
	\resizebox{\textwidth}{!}{
		\begin{tabular}{lccccccccc}
			\hline
			\textbf{Case} & \textbf{Accuracy} & \textbf{Precision} & \textbf{Recall} & \textbf{F1} & \textbf{MCC} & \textbf{Balanced Acc} & \textbf{ROC-AUC} & \textbf{LogLoss} & \textbf{Brier} \\
			\hline
			Clean\_Base  & 0.9960 & 0.9961 & 0.9960 & 0.9960 & 0.9956 & 0.9960 & 0.99996 & 1.4693 & 0.6582 \\
			Clean\_Def   & 0.9720 & 0.9722 & 0.9720 & 0.9719 & 0.9689 & 0.9720 & 0.9991 & 1.5132 & 0.6734 \\
			Robust\_Base & 0.6920 & 0.7711 & 0.6920 & 0.6864 & 0.6663 & 0.6920 & 0.9912 & 1.6933 & 0.7259 \\
			Robust\_Def  & 0.7270 & 0.7966 & 0.7270 & 0.7346 & 0.7047 & 0.7270 & 0.9889 & 1.6817 & 0.7236 \\
			\hline
		\end{tabular}
	}
	\caption{Performance comparison on clean and adversarial data after optimized $k$ value for NNMF $(k = 70)$.}
	\label{tab:results_full_optimized}
\end{table*}

\section{Conclusion}

{This work represents an initial investigation toward extending the proposed method to multi-class settings, with further evaluation on larger benchmarks planned as future work.}

\section*{Acknowledgement}

This work was supported by the Distinguished Professor Program of \'{O}buda University. The authors are also grateful for the possibility of using the HUN-REN Cloud \url{https://science-cloud.hu/en}~\cite{H_der_2022} which helped us achieve some particular results published in this paper. Hiba Adil Al-kharsan gratefully acknowledges the financial support of the Stipendium Hungaricum Doctoral Program, managed by the Tempus Public Foundation.

\bibliographystyle{plain}
\bibliography{references}

\end{document}

%% file: cscsstyle.tex
\documentclass[11pt,twoside]{article}
\usepackage{palatino}
\usepackage{graphicx}
\usepackage{amsmath, amssymb, amsbsy}
\usepackage[latin2]{inputenc}
\usepackage{t1enc}

\usepackage[mathcal]{eucal}

\usepackage{amsthm}

\usepackage[mathcal]{eucal}
\usepackage{array}

\def\contradict{\mathrel=\kern-2pt\mathrel|}

\usepackage[T1]{fontenc}

\usepackage{program}
\usepackage{alltt}
\usepackage{tabularx}

\newcommand{\Title}[1]{\setcounter{footnote}{0} {\begin{center}\Large\noindent \textbf{ #1}\end{center}}}
\newcommand{\Author}[1]{{\begin{center} \par\noindent\textbf{ #1}\end{center}}}

\newenvironment{Abstract}{\vspace*{0.em}\par\noindent\textbf{Abstract:~}}{}
\def\R{ \hbox{I\kern-.1667em R}}

\def\thebibliography#1{\subsubsection*{References}\list
 {[\arabic{enumi}]}{\settowidth\labelwidth{[#1]}\leftmargin\labelwidth
 \advance\leftmargin\labelsep
 \usecounter{enumi}}
 \def\newblock{\hskip .11em plus .33em minus .07em}
 \sloppy\clubpenalty4000\widowpenalty4000
 \sfcode`\.=1000\relax}

\def\Name1#1{{\vspace{0.5mm}\rm  #1:\par \vspace{1mm}}}

    {\begin{list}{\hbox{}}%
           {
       \setlength{\parsep}{0pt}
       \setlength{\itemsep}{0em}}}%
    {\end{list}}